\documentclass[conference,table]{IEEEtran}
\IEEEoverridecommandlockouts
\usepackage{cite}
\usepackage{amsmath,amssymb,amsfonts}
\usepackage{algorithmic}
\usepackage{graphicx}
\usepackage{textcomp}
\usepackage{xcolor}
\usepackage{multirow}
\usepackage{siunitx}
\usepackage{booktabs}
\usepackage{color,soul}
\usepackage{eso-pic}

\usepackage[hidelinks]{hyperref} 
\usepackage{orcidlink} 

\definecolor{darkgreen}{HTML}{006400} 

\AddToShipoutPictureBG{

  \AtPageCenter{%
    \makebox(0,0){\rotatebox{45}{%
      \scalebox{6}{%
        \textnormal{\textcolor{gray!30}{PREPRINT}}%
      }%
    }}%
  }%
  
    \AtPageUpperLeft{%
      \hspace{\paperwidth/2-6cm}%
      \raisebox{-1.5cm}{%
        \textnormal{\textcolor{gray}{This is a preprint. For the final version, please refer to the conference proceedings.}}%
      }%
    }%
}

\def\BibTeX{{\rm B\kern-.05em{\sc i\kern-.025em b}\kern-.08em
    T\kern-.1667em\lower.7ex\hbox{E}\kern-.125emX}}
\begin{document}

\title{Limitations of Physics-Informed Neural Networks: a Study on Smart Grid Surrogation}


\author{
\IEEEauthorblockN{Julen Cestero Portu\IEEEauthorrefmark{1}\IEEEauthorrefmark{2}\orcidlink{0000-0002-6670-6255},
Carmine Delle Femine\IEEEauthorrefmark{1}\orcidlink{0009-0005-7964-8911},
Kenji S. Muro\IEEEauthorrefmark{1}\orcidlink{0009-0003-3135-8988},
Marco Quartulli\IEEEauthorrefmark{1}\orcidlink{0000-0001-5735-2072} and
Marcello Restelli\IEEEauthorrefmark{2}\orcidlink{0000-0002-6322-1076}}
\IEEEauthorblockA{\IEEEauthorrefmark{1}Department of Energy and Environment\\
Vicomtech, Donostia - San Sebastián, Gipuzkoa, Spain\\
Corresponding Author: jcestero@vicomtech.org}
\IEEEauthorblockA{\IEEEauthorrefmark{2}Department of Electronics, Information and Bioengineering\\
Politecnico di Milano, Milano, Italy}}

\maketitle

\begin{abstract}
Physics-Informed Neural Networks (PINNs) present a transformative approach for smart grid modeling by integrating physical laws directly into learning frameworks, addressing critical challenges of data scarcity and physical consistency in conventional data-driven methods. This paper evaluates PINNs' capabilities as surrogate models for smart grid dynamics, comparing their performance against XGBoost, Random Forest, and Linear Regression across three key experiments:  interpolation, cross-validation, and episodic trajectory prediction. By training PINNs exclusively through physics-based loss functions—enforcing power balance, operational constraints, and grid stability—we demonstrate their superior generalization, outperforming data-driven models in error reduction. Notably, PINNs maintain comparatively lower MAE in dynamic grid operations, reliably capturing state transitions in both random and expert-driven control scenarios, while traditional models exhibit erratic performance. Despite slight degradation in extreme operational regimes, PINNs consistently enforce physical feasibility, proving vital for safety-critical applications. Our results contribute to establishing PINNs as a paradigm-shifting tool for smart grid surrogation, bridging data-driven flexibility with first-principles rigor. This work advances real-time grid control and scalable digital twins, emphasizing the necessity of physics-aware architectures in mission-critical energy systems.
\end{abstract}


\section{Introduction}


Deep Learning has brought a significant leap in feature learning that has caught the interest of both industry and academia, as it can be used for new data-driven solutions in which complete physical properties are not fully understood \cite{pinns4pspaper}. This is prominent in the case of energy systems such as smart grids, in which the physical conditions of each component are well understood, but the possible interactions that they can have between each other are not \cite{pinnssgnapa}. However, Deep Neural Networks still have difficulties making predictions of real systems, mainly due to the need for large quantities of good quality data and the possibility of providing physically unfeasible results \cite{pinns4pspaper}.

However, Physics-Informed Neural Networks (PINNs) can have the physical equations that define the system explicitly included in their loss function, and thus, the training phase has a physical guidance that accelerates the training process, reduces computational needs, and provides reliable solutions \cite{pinns4pspaper}. This is why PINNs are well suited for smart grid operations to overcome the challenges related to computational needs and the amount of input data required due to their highly complex nature \cite{sgdimension}.

Control of smart grids is a complex topic due to the nature of electrical grids and the need to coordinate equipment with different behaviors and necessities while maintaining a high degree of safety \cite{rlisgud}. Models and algorithms need to be able to predict and tackle emergency situations, which might be outside of training datasets in case machine learning strategies are used. In these cases, PINNs can also be helpful as they have greater generalization capabilities, so they can provide physically reliable results outside the training dataset, which in smart grid operation could involve outlying or emergency situations that can rarely be experienced. Similarly, the interpolation capabilities of PINNs are also interesting for providing solutions for specific variables or conditions where measurement is difficult \cite{sgextrapinterp}.

In this work, we focus on evaluating the capabilities of PINNs in terms of interpolation, extrapolation, and feasibility as a replacement for an episodic simulator (surrogate model of Reinforcement Learning environments) applying them into a smart grid simulation environment called Gym-ANM \cite{anm6paper}. For that, we compare the performance of PINNs against other state-of-the-art models such as XGBoost, Linear Regression (LR), and Random Forest (RF) in three different experiments. Specifically, we study the capabilities of all these models trained in different datasets and validate them against challenging scenarios, such as smart grid management. We find that PINNs outperform all the other models in terms of generalization and error minimization, although XGBoost can defy its performance in specific experiments. Finally, we see that in episodic trajectories, the PINN model is the only one capable of maintaining good accuracy across all the steps, being a suitable candidate for providing faster interaction with the grid and allowing optimization algorithms in real-time grid control systems.

\section{Related work}

\subsection{Application of PINNs in smart grids}

Although new advances in machine learning and deep learning have brought a leap in numerous tasks in power systems, there are still some challenges present, such as a need for large amounts of high-quality training data, solutions incompatible with physics, or low capabilities of generalization and interpretability of the results \cite{dlprobsinsg}. PINNs can address such challenges due to their inclusion of the physical principals in the loss function \cite{pinns4psdynamics}.

In power systems and smart grids in particular, the use of PINNs has also caught attention in the literature for their interest in topics such as estimation of the state parameters of an electrical grid \cite{gridparams}, dynamic regulation of electrical parameters \cite{dr1, pinns4psdynamics} including energy storage characteristics~\cite{dr2}.

A particularly promising application of PINNs in power systems, such as smart grids, lies in load flow calculation and Optimal Power Flow. Although conventional OPF solvers may provide guaranteed optimality given sufficient time, the computational speed of PINNs, combined with their high accuracy and physical consistency, makes them a compelling alternative for time-critical applications such as real-time grid management and control, where reliable and rapid solutions are essential \cite{pinns4pspaper}. Recent research has demonstrated this capability by incorporating physical constraints into the loss function of the PINNs. Examples of such constraints are Kirchoff's laws \cite{pinn4lf1}, the physical limits of the grid \cite{kktstealers2}, and the Karush-Kuhn-Tucker optimality conditions \cite{kktcarmine, kktstealers}.

The advantages offered by PINNs can be further amplified when combined with other algorithms, such as Reinforcement Learning (RL), to design reliable control systems that take into account the physical constraints of the grid itself, ensure operational safety, and optimize the management of grid assets, including photovoltaic (PV) panels and battery energy storage systems, to enhance self-consumption and promote sustainability \cite{rlisgud}. Gym-ANM is an example of such an RL environment, which has the physical properties of its components explicitly defined \cite{anm6paper}. A surrogate model of such an environment enhanced by PINNs can combine all of its advantages, bringing lower computational needs, an objective-focused control, and reliable solutions \cite{papersuiza}.


\subsection{Extrapolation and generalization}


In the field of generalization capabilities of PINNs, solving Partial Differential Equations (PDEs) is the topic with the most attention in the literature, in which several studies were capable of obtaining accurate solutions outside of the training dataset in the vicinity of the dataset boundaries and with innovative regularization techniques to reduce extrapolation errors \cite{generalisationofpinns, advancinggen, lomismoperoLS}. Promising extrapolation capabilities have been observed with physical equations such as the advection equation, compressible Euler equations, Schrödinger equation, or the Poisson equation, and different extrapolation methods have been proposed depending on the nature of the physical phenomenon \cite{moregen, pinnextrap1, pinnextrap2}. In a more realistic scenario, a PINN successfully predicted fatigue cracks in gas turbine nozzles with limited training data, providing accurate results even outside the training domain \cite{pinns4cracks}.

These extrapolation capabilities that PINNs can offer in differential systems may also be applicable in nondifferential ones but with high complexity, such as in smart grids, and thus there is an interest in testing PINNs extrapolation performance applied to smart grid simulation environments such as Gym-ANM \cite{anm6paper}.

\subsection{Interpolation}

PINNs rely on the domain of their underlying physical equations. Therefore, solutions can be calculated at any point of the domain. As highlighted previously, real-world smart grid operation often faces spatial and temporal constraints in measurement acquisition—a critical challenge where physics-informed interpolation enables reliable state estimation in uninstrumented nodes or non-observable grid regions \cite{apinn}.

Regarding this, we made these findings in the literature, for example, in physical equations such as the Schrödinger and Burgers' equation in which there were using initial data at the boundaries of each dimension, surrogate models based on PINNs were successful at predicting solutions in the interior space between the boundaries \cite{pinns4pde}. PINNs were also applied in a simulator of a hydraulic subsurface transport network to facilitate the calculation of hydraulic state parameters across the transport network from a few measurement points \cite{pinns4physics}. PINNs were also used for functional interpolation applied to stiff chemical kinetic problems \cite{paperfunctionalinterpol}.

Although these examples were mostly applied in differential equations, they may hold for nondifferential systems as well, such as smart grids.


\section{Methods}


In light of this, we developed a Physics-Informed Neural Network (PINN) framework tailored for the surrogation of smart grid models, in particular the one of ANM6-Easy, integrating on their loss function the information of load flow calculation and efficient grid operation among others, testing the PINNs' effects on simulation speed and accuracy.

To evaluate its performance, three key experiments were conducted, focusing on interpolation, cross-validation and episodic trajectory prediction. All experiments compared the PINN against traditional data-driven models such as XGBoost, Random Forest, and Linear Regression. These experiments were designed to rigorously test the model's generalization, physical consistency, and reliability in dynamic grid operations.

\subsection{Case study and datasets}

The case study utilizes the ANM6-Easy environment from Gym-ANM, which simulates an active distribution network. The operational state of this network at any given time step is defined by a set of state variables representing the grid's physical condition. For the surrogate modeling task evaluated in this paper, these state variables include: bus voltage magnitudes, bus voltage phase angles, active power injections from controllable generators, reactive power injections from controllable generators, and the state-of-charge (SOC) along with active/reactive power injections/withdrawals for the distributed energy storage (DES) units. The surrogate models developed and compared in this study aim to predict these state variables for the subsequent time step, given the current state and the control actions applied.

The PINN is trained exclusively using a physics-based loss function, ensuring that its predictions remain physically consistent while avoiding reliance on empirical data loss. This methodology allows us to assess whether the model can generalize beyond the distribution of its training set and accurately predict system states outside its training domain.

To this end, we propose a structured state-space partitioning approach that isolates different subdomains of the system’s dynamics. Specifically, the state variables (excluding auxiliary variables and the slack generator's power injections) are discretized into 8 separate bins, each defining a distinct dataset for model training.

Given the bounded nature of these variables, we construct 8 disjoint datasets by partitioning each state variable into equal-width bins spanning its observed range. The n-th dataset is then defined as the subset of system states where every variable belongs to its corresponding n-th bin. This results in 8 datasets, each representing a distinct region of the state space with non-overlapping subdomains.

\subsection{Experiments definition}

\subsubsection{Interpolation}\label{method-ex2}

Each model was trained on the entire domain and later evaluated on the subspaces.

\subsubsection{Cross-validation}\label{method-ex3}

Inspired by the work of Cestero et al. \cite{papersuiza}, we tried to replicate the evaluation in our use case. Specifically, since these models try to surrogate a simulator that can be used as a RL environment, we create a dataset using trajectories of agents using a random policy and an expert policy (a policy from an already trained agent that converged to an optimal solution). We can make use of our dataset containing samples from the whole state space and, following the naming conventions from Cestero, call it the `Generative' dataset and the dataset with the trajectories of agents `Agent-based' dataset. With these datasets we retrain our models. Our PINN model makes an exception, since it has no need of datasets for training, only an internal sampling process similar to the so-called Generative methods. With these new models, we are able to cross-validate the performance of the models trained in the whole state space (now called Generative dataset) against the models trained in the agent-based dataset and vice versa.

\subsubsection{Episodic accuracy}\label{method-ex4}

We tried to evaluate the effectiveness of the models in a more realistic scenario. For that, and taking advantage of the nature of the simulator to be used as an RL environment, we try to simulate two episodes, one performed by an expert agent and another one performed by a random agent, and we calculate in each step the Mean Absolute Error (MAE) of each model, comparing them to the real transitions from the simulator.
\subsection{PINN architecture}
The PINN used in this study is structured into modular sub-networks, each specialized for different components of the smart grid:
\begin{itemize}
\item Voltage prediction module: Estimates bus voltage magnitudes and phases.
\item Generator control module: Models active and reactive power outputs of non-slack generators, using residual blocks to facilitate gradient flow.
\item Storage system module: Captures the state transitions and power injections of distributed energy storage (DES) units.
\end{itemize}
Each module consists of fully connected layers with LeakyReLU activations, and residual connections are used to improve stability during training. The training was performed by sampling a Sobol sequence of subspace points at each step and calculating the physical loss function in them.

Training is performed using a purely physics-based loss function without reliance on empirical data. The loss function enforces:
\begin{itemize}
\item Power balance constraints, ensuring conservation of energy.
\item Voltage and power flow equations, maintaining system stability.
\item Device-specific operational limits, such as generation caps and storage constraints.
\end{itemize}
Training batches are generated using Sobol sequences, ensuring uniform coverage of the input space. Early stopping is applied to monitor convergence, preventing overfitting. The model is trained using adaptive learning rate schedules and optimized through AdamW algorithm.
\section{Results}
    In this section, we evaluate the performance of the PINN against a set of reference models (i.e., Linear Regressor, Random Forest, and XGBoost). For that, we test the models against the previously defined experiments and compare their accuracy in terms of different error metrics. 
    
    In general, we compare the models in terms of Mean Squared Error (MSE) and Coefficient of determination (R$^2$). We find that some models perform similarly to the PINN model in most experiments, but none are on par with it in terms of MSE.
    
\subsection{Experiment 1: Interpolation}

    \begin{figure}[ht]
      \centering
      \includegraphics[width=\linewidth]{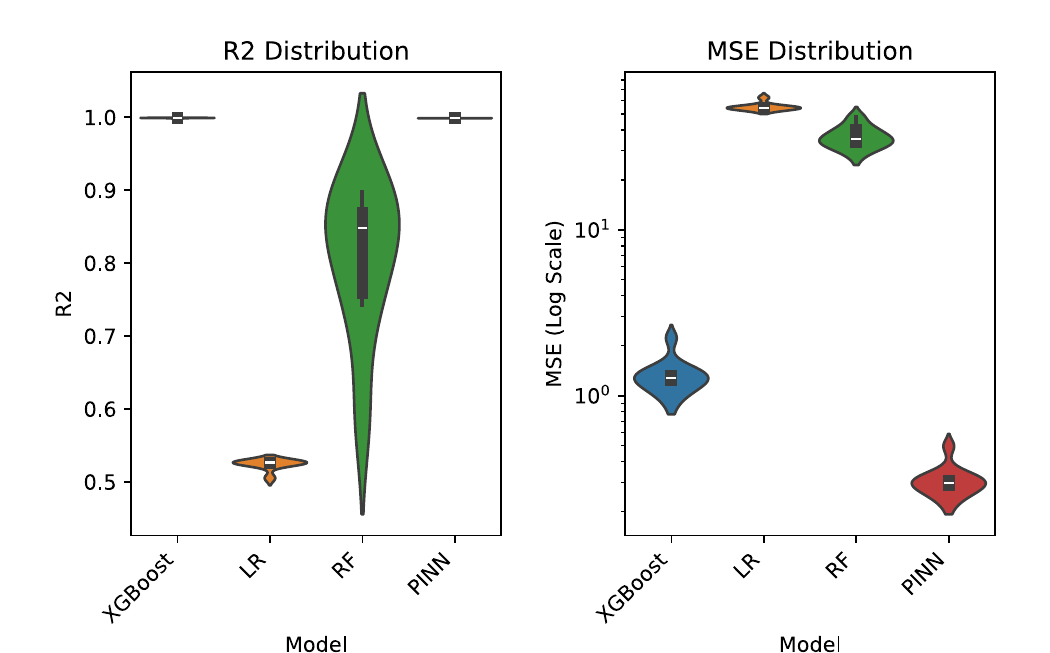}
      \caption{Distribution of the performance in terms of R$^2$ and MSE of the studied models trained in the whole state space and tested in all the considered sub-datasets. The results clearly show that, even though XGBoost and PINN models show similar performances in terms of R$^2$, the PINN model clearly outperforms the XGBoost model in terms of MSE. On the other hand, LR shows insufficient performance, while RF shows variable performance across the datasets.}
      \label{fig:exp2}
    \end{figure}
    
    In Experiment 1, we evaluate the performance of the models when they are trained using a dataset containing the whole state space but tested in the different subsets of the space.

\begin{table}[htbp]
    \centering
    \caption{Model Interpolation Scores}
    \label{tab:interpolation}
    \small
    \begin{tabular}{llcc}
        \toprule
        \textbf{Model} & \textbf{Dataset} & \textbf{R2} & \textbf{MSE} \\
        \midrule
        \multirow{9}{*}{XGBoost} 
         & D1 & 0.9991 & 2.2262 \\
         & D2 & 0.9991 & 1.3002 \\
         & D3 & 0.9990 & 1.2879 \\
         & D4 & 0.9991 & 1.2207 \\
         & D5 & 0.9990 & 1.2273 \\
         & D6 & 0.9990 & 1.2421 \\
         & D7 & 0.9990 & 1.2356 \\
         & D8 & 0.9989 & 1.3523 \\
         & \textbf{Average} & \textbf{\textcolor{darkgreen}{0.9989}} & \textbf{1.3915} \\
        \midrule
        \multirow{9}{*}{Linear Regressor} 
         & D1 & 0.5049 & 62.6855 \\
         & D2 & 0.5257 & 54.6559 \\
         & D3 & 0.5262 & 54.3356 \\
         & D4 & 0.5264 & 53.8231 \\
         & D5 & 0.5267 & 54.0813 \\
         & D6 & 0.5263 & 54.3364 \\
         & D7 & 0.5262 & 54.2558 \\
         & D8 & 0.5258 & 54.4023 \\
         & \textbf{Average} & \textbf{\textcolor{red}{0.5235}} & \textbf{\textcolor{red}{55.3220}} \\
        \midrule
        \multirow{9}{*}{Random Forest} 
         & D1 & 0.5913 & 47.8471 \\
         & D2 & 0.8968 & 31.8090 \\
         & D3 & 0.8604 & 34.2322 \\
         & D4 & 0.7642 & 39.6275 \\
         & D5 & 0.7421 & 41.2212 \\
         & D6 & 0.8499 & 34.0764 \\
         & D7 & 0.8935 & 32.1320 \\
         & D8 & 0.8461 & 36.1481 \\
         & \textbf{Average} & \textbf{0.8055} & \textbf{37.1404} \\
        \midrule
        \multirow{9}{*}{PINN} 
         & D1 & 0.9981 & 0.4955 \\
         & D2 & 0.9986 & 0.3036 \\
         & D3 & 0.9986 & 0.2922 \\
         & D4 & 0.9986 & 0.2862 \\
         & D5 & 0.9986 & 0.2892 \\
         & D6 & 0.9986 & 0.3041 \\
         & D7 & 0.9986 & 0.2900 \\
         & D8 & 0.9986 & 0.3080 \\
         & \textbf{Average} & \textbf{0.9985} & \textbf{\textcolor{darkgreen}{0.3211}} \\
        \bottomrule
    \end{tabular}
\end{table}

    Figure \ref{fig:exp2} shows the distribution of the results for the different models. This comparative analysis reveals distinct patterns in predictive accuracy and error minimization. As illustrated by the Figure, plot distributions of MSE and R$^2$, XGBoost, and PINNs emerge as the top performers, albeit with nuanced strengths. XGBoost demonstrates a slight edge in R$^2$ scores, indicating superior explanatory power across diverse data subsets, while PINNs achieve markedly lower MSE values, underscoring their precision in minimizing prediction errors. However, both methods show promising results, although PINNs’ integration of physical constraints enhances error control in a superior way to XGBoost, as depicted in their difference of MSE. In contrast, Linear Regression (LR) exhibits consistently poor performance, highlighting its inadequacy for nonlinear or high-dimensional patterns. Random Forest (RF) displays variable reliability, likely due to sensitivity to dataset-specific noise. These findings advocate for context-driven model selection, prioritizing XGBoost for variance explanation and PINNs for error-critical applications while emphasizing the limitations of linear approaches in complex regression tasks. Specific results for each of the defined datasets are depicted in Table \ref{tab:interpolation}.

\subsection{Experiment 2: Cross-validation}

    As explained in \ref{method-ex3}, a PINN and reference models were trained.
\begin{table*}[htbp]
    \centering
    \caption{Model Performance Cross-validation}
    \label{tab:x-val}
    \small
    \begin{tabular}{@{}llcc@{\hspace{2em}}cc@{\hspace{2em}}cc@{}}
        \toprule
        \multirow{2}{*}{\textbf{Training Method}} & \multirow{2}{*}{\textbf{Model}} & \multicolumn{2}{c}{\textbf{Generative Testing}} & \multicolumn{2}{c}{\textbf{Agent-Based Testing}} & \multicolumn{2}{c}{\textbf{Average Results}} \\
        \cmidrule(lr){3-4} \cmidrule(lr){5-6} \cmidrule(l){7-8}
         & & R² & MSE & R² & MSE & R² & MSE \\
        \midrule
        \multirow{3}{*}{Generative} 
         & RF & 0.9701 & 11.6084 & 0.8425 & 52.8641 & \textbf{0.9063} & \textbf{32.2363} \\
         & LR & 0.5237 & 56.3810 & 0.4870 & {56.5669} & \textbf{0.5054} & \textbf{56.4739} \\
         & XGB & {0.9997} & 0.5259 & 0.9916 & 6.4557 & \underline{\textbf{0.9957}} & \underline{\textbf{3.4908}} \\
        
        \addlinespace[0.5em]
        \multirow{1}{*}{Other} 
         & PINN & 0.9985 & {0.4106} & 0.9975 & 0.5410 & \textcolor{darkgreen}{\textbf{0.9980}} & \textcolor{darkgreen}{\textbf{0.4758}} \\
        
        \addlinespace[0.5em]
        \multirow{3}{*}{Agent-Based} 
         & XGB & 0.0441 & 92.8956 & {0.9997} & {0.2764} & \textbf{0.5219} & \textbf{46.5860} \\
         & LR & {-211.8839} & {30128.5165} & 0.9421 & 15.7468 & \textcolor{red}{\textbf{-58.4709}} & \textcolor{red}{\textbf{15072.1317}} \\
         & RF & 0.5234 & 100.9040 & 0.9700 & 7.6132 & \textbf{0.7467} & \textbf{54.2586} \\
        \bottomrule
    \end{tabular}
\end{table*}

    The results of this comparison can be seen in Table \ref{tab:x-val}. These results demonstrate a critical dependence of model performance on the alignment between training methodology and testing paradigm. Models trained via generative methods (RF, LR, XGB) exhibited strong performance in generative testing but showed reduced efficacy in agent-based testing, highlighting domain-specific robustness. On the other hand, agent-based trained models achieved exceptional performance in agent-based testing but suffered significant degradation in generative environments. Notably, the PINN model demonstrated exceptional cross-paradigm generalization, suggesting physics-informed approaches may transcend methodology biases. The catastrophic failure of agent-trained LR in generative testing reveals fundamental limitations in linear models for cross-domain extrapolation. We underline the results of the XGB model since their results are close to the results of the PINN model, albeit slightly worse.

\subsection{Experiment 3: Episodic accuracy}

As outlined in \ref{method-ex4}, two episodes were simulated, respectively using an expert and a random agent, and hence the models were evaluated on them.
\begin{figure*}[t]
  \centering
  \includegraphics[width=\linewidth]{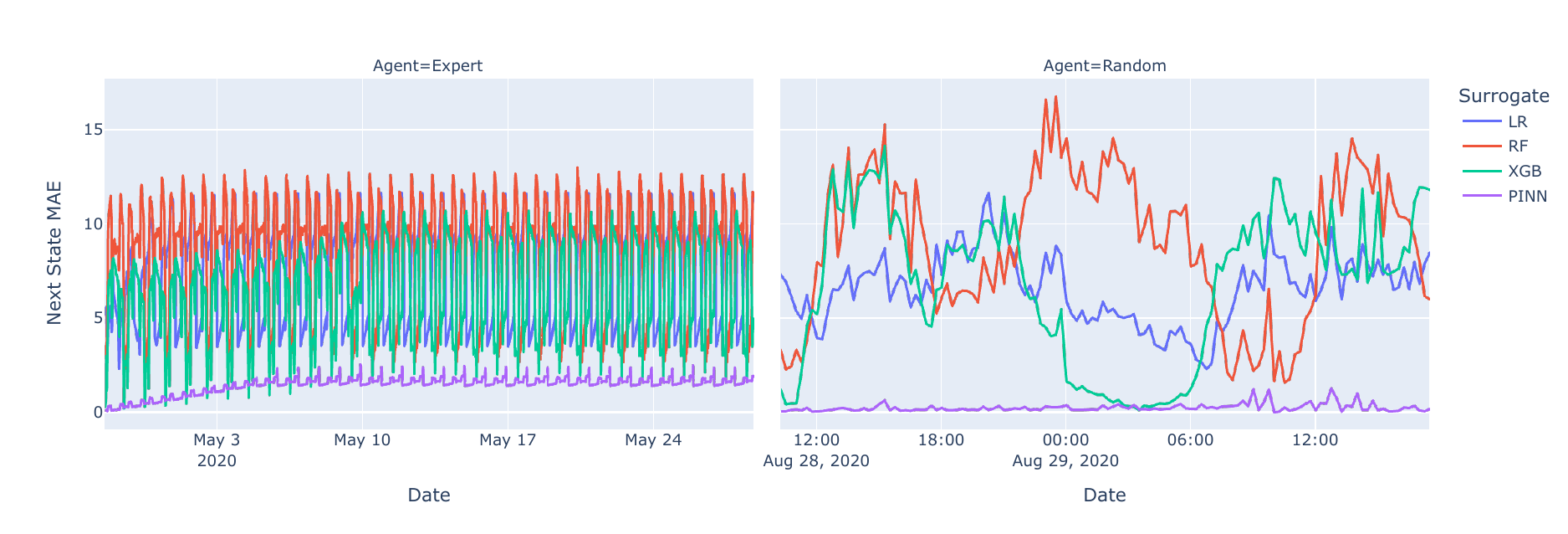}
  \caption{Step-by-step comparison of the error (MAE) of the tested models over two different episodes: an episode driven by an optimal policy trained with RL algorithms (Expert), and an episode driven by a random agent (Random). Both episodes underline the effectiveness of PINNs to cover accurately the state transitions of the environment due to its capability to understand the physical nature of the problem, while this nature hinders the performance of the other models. Although in some states, some models, notably XGBoost, show almost the best accuracy over all the models, it varies over the episode, showing their limitations.}
  \label{fig:episodic_comparison}
\end{figure*}
Figure \ref{fig:episodic_comparison} shows the result of this experiment. The expert-trained agent exhibits a characteristic daily operational frequency emerging from its learned policy to balance charge cycles. This behavior manifests in two distinct simulation phases: (1) an initial ramp-up period characterized by rapid charge accumulation, followed by (2) a stability phase maintaining state-of-charge (SOC) within optimal thresholds (20\%-80\%). This bimodal pattern directly reflects the battery's intrinsic dynamics—gradual energy storage until achieving sufficient capacity to sustain controlled oscillations between SOC boundaries.

Our physics-informed neural network (PINN) uniquely captures these transitions with comparative error, while conventional models demonstrate substantial performance limitations. Although XGBoost achieves peak accuracy in specific states of the random episode, its predictive reliability degrades significantly in alternative operational regimes. Notably, RF underperforms LR in both episodes, contrary to theoretical expectations from prior benchmark analyses. All non-PINN models exhibit state-dependent accuracy fluctuations, rendering them unsuitable for precision-critical RL environments requiring stepwise reliability.

The erratic performance in the random episode further accentuates PINN's methodological superiority—while baseline models alternate between high and low accuracy depending on SOC conditions, PINN maintains stable precision across all operational phases. This divergence suggests PINN successfully encodes first-principles battery physics, as well as other internal smart grid mechanisms, enabling robust generalization beyond training distributions. In contrast, other architectures appear limited to mimicking dataset-specific transitions without capturing the underlying grid's energy flows. These results underscore that while data-driven models achieve high nominal performance metrics, their inability to maintain precision under dynamic real-world conditions necessitates physics-aware architectures for mission-critical applications.

\section{Conclusions}

This study rigorously evaluates the capabilities and limitations of PINNs in modeling smart grid dynamics, with a focus on their interpolation, extrapolation, and generalization performance compared to traditional data-driven methods. Our findings demonstrate that PINNs achieve superior physical consistency and error minimization, even in out-of-distribution scenarios, while maintaining computational efficiency. 

PINNs significantly outperform conventional models (XGBoost, RF, LR) in cross-validation tasks, achieving an average low MSE  across generative and agent-based datasets. Their physics-informed loss function enables reliable predictions beyond training domains, which is critical for safety-critical grid operations. Therefore, we verified that PINNs are the most capable studied model to generalize the information acquired during the training.

Another key point of PINNs is their capability to interpolate data. While XGBoost exhibits competitive R$^2$ scores, PINNs reduce prediction errors,  represented by the MAE metric, by almost 80\% in interpolation tasks, proving their ability to enforce physical constraints without empirical data reliance.

One of the most critical points of this study is the episodic reliability of the models. In dynamic simulations, PINNs maintain comparatively smaller MAE across both random and expert-driven grid operations, whereas data-driven models show erratic performance. This stability underscores PINNs’ suitability for RL environments requiring stepwise precision.

However, we also found some limitations to the PINNs. They exhibit slight performance degradation in extreme operational regimes (e.g., very high/low state-of-charge), highlighting challenges in modeling stiff system transitions. Additionally, their training requires careful balancing of physical loss terms—a non-trivial design consideration. This and designing the architecture of PINN models could be tasks that require very deep domain-specific knowledge, which may be hard to replicate if the information for the aimed scenario is lacking or obscure. These reasons may be enough for some authors to use any other more data-driven approaches. However, in some cases, these data-driven methods have been proven that they are not enough to represent thoroughly the physics of RL environments.

These results advance smart grid surrogation by demonstrating that PINNs bridge the gap between data-driven flexibility and physics-based rigor. Future work will focus on integrating PINNs as surrogates of RL environments, determining their capabilities to maximize the performance of computationally expensive simulation tasks, and bridging the gap between the simulation paradigm and optimization algorithms, such as RL. This approach holds promise for real-time grid control and scalable digital twin development.

\section*{Funding}

This research was funded by the European Union’s Horizon Europe research and innovation program under HORIZON-EUSPA-2021-SPACE call, grant agreement No. 101082355, with
the acronym RESPONDENT

\end{document}